\documentclass[10pt,twocolumn,letterpaper,pagebackref,breaklinks,colorlinks]{article}

\usepackage[pagenumbers]{wacv} 

\usepackage{graphicx}
\usepackage{amsmath}
\usepackage{amssymb}
\usepackage{booktabs}
\usepackage{epsfig}
\usepackage{graphicx}
\usepackage{amsmath}
\usepackage{amssymb}
\usepackage{xfrac}
\usepackage[mathscr]{euscript}
\usepackage{appendix}
\usepackage{algorithm}
\usepackage{algpseudocode}
\usepackage{cancel}
\usepackage{caption} 
\usepackage{color,soul}
\usepackage{tabularx}
\usepackage{subcaption}
\usepackage{mathtools}
\usepackage{paralist} 
\usepackage{colortbl}
\usepackage{multirow}
\usepackage{graphicx}
\usepackage{booktabs}
\usepackage[algo2e]{algorithm2e}
\usepackage{amssymb}
\usepackage{pifont}
\usepackage{orcidlink}
\usepackage{varwidth}
\usepackage{wrapfig,lipsum,booktabs}

\usepackage[accsupp]{axessibility}  

\definecolor{yellow}{rgb}{1, 1, 0.7}
\definecolor{orange}{rgb}{1, 0.85, 0.7}
\definecolor{red}{rgb}{1, 0.7, 0.7}
\definecolor{lightyellow}{rgb}{1,1, 0.8}

\definecolor{wincolor}{rgb}{0.85, 0.0, 0.0}

\definecolor{darkyellow}{rgb}{0.8, 0.8, 0.5}
\definecolor{darkred}{rgb}{0.7, 0.3, 0.3}
\definecolor{darkgreen}{rgb}{0.3, 0.7, 0.3}
\definecolor{blue}{rgb}{0, 0, 1.0}
\definecolor{green}{rgb}{0, 1.0, 0}
\definecolor{pink}{rgb}{1, 0.4, 0.7}

\newcommand{\Mat}{\boldsymbol}

\usepackage[capitalize]{cleveref}
\crefname{section}{Sec.}{Secs.}
\Crefname{section}{Section}{Sections}
\Crefname{table}{Table}{Tables}
\crefname{table}{Tab.}{Tabs.}

\def\wacvPaperID{2155} 
\def\confName{WACV}
\def\confYear{2025}

\begin{document}

\title{GANESH: Generalizable NeRF for Lensless Imaging}

\author{Rakesh Raj Madavan$^1$\thanks{Equal Contribution.} , \; Akshat Kaimal$^1$\footnotemark[1] , \; Badhrinarayanan K V$^1$\footnotemark[1] , \; Vinayak Gupta$^2$
\\
Rohit Choudhary$^2$, \; Chandrakala Shanmuganathan$^1$, \; Kaushik Mitra$^2$ \\
$^1$Shiv Nadar University, Chennai\;\;\;\; $^2$Indian Institute of Technology, Madras 
}
\maketitle

\begin{abstract}
   Lensless imaging offers a significant opportunity to develop ultra-compact cameras by removing the conventional bulky lens system. However, without a focusing element, the sensor's output is no longer a direct image but a complex multiplexed scene representation. Traditional methods have attempted to address this challenge by employing learnable inversions and refinement models, but these methods are primarily designed for 2D reconstruction and do not generalize well to 3D reconstruction. We introduce GANESH, a novel framework designed to enable simultaneous refinement and novel view synthesis from multi-view lensless images. Unlike existing methods that require scene-specific training, our approach supports on-the-fly inference without retraining on each scene. Moreover, our framework allows us to tune our model to specific scenes, enhancing the rendering and refinement quality. To facilitate research in this area, we also present the first multi-view lensless dataset, LenslessScenes. Extensive experiments demonstrate that our method outperforms current approaches in reconstruction accuracy and refinement quality. Code and video results are available at \href{https://rakesh-123-cryp.github.io/Rakesh.github.io/}{https://rakesh-123-cryp.github.io/Rakesh.github.io/}
\end{abstract}


\section{Introduction}
\label{sec:intro}


In recent years, mask-based lensless imaging systems have received significant interest due to their potential to offer compact, lightweight, and cost-efficient alternatives to traditional cameras\cite{khan2020flatnet}. Rather than utilizing standard optical lenses, these systems rely on amplitude\cite{asif2016flatcam} or phase masks\cite{boominathan2020phlatcam, antipa2017diffusercam} placed in close proximity to the sensor. This design not only minimizes the physical dimensions and mass of the imaging device, i.e., a smaller form factor, but also enables the use of non-traditional sensor geometries, such as spherical, cylindrical, or even flexible configurations\cite{tremblay2007ultrathin}. In the absence of a conventional focusing element, the measurements captured by the sensor are not straightforward images of the scene. Instead, they consist of intricate, multiplexed data that encode the light information from the scene in a highly compressed and non-intuitive form. This image formation model necessitates advanced computational techniques to decode and reconstruct the original scene, as the sensor no longer produces a one-to-one representation of the visual information but rather a superimposition of light intensities across the entire field of view. 

\begin{figure}
  \includegraphics[width=0.47\textwidth]{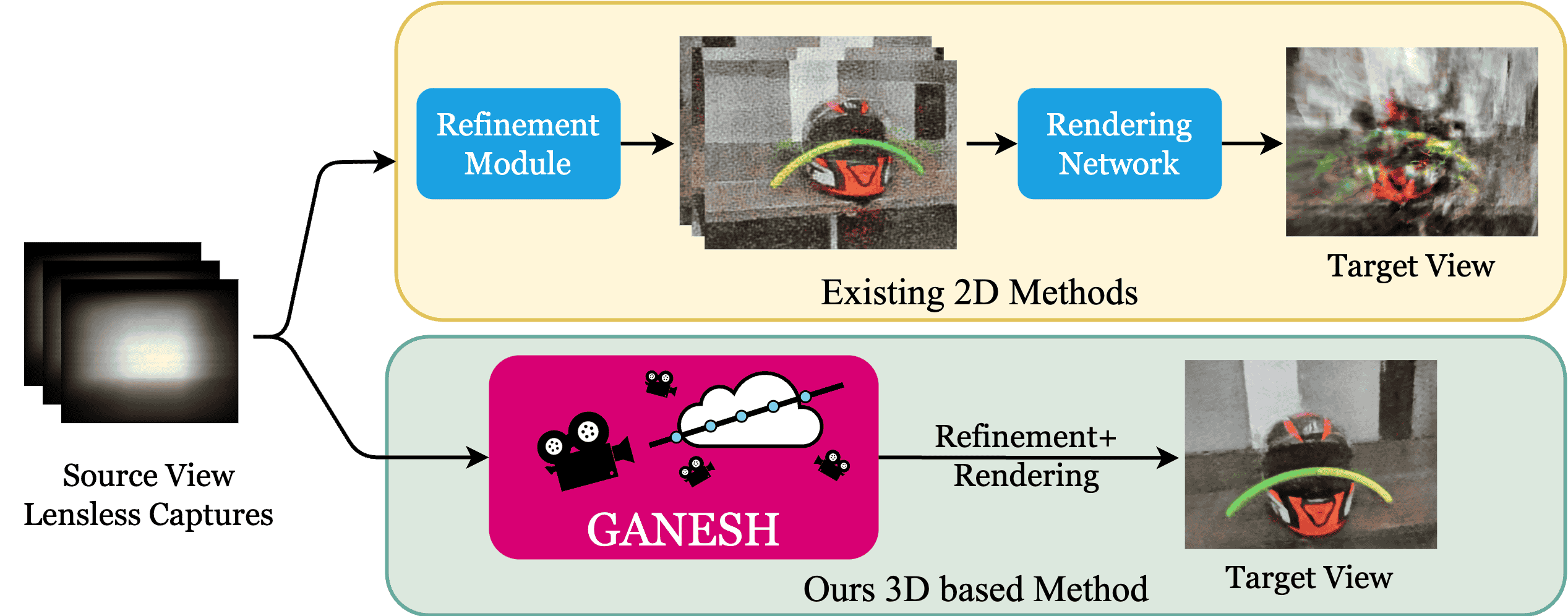}
  \caption{Reconstructing 3D scenes from multi-view lensless captures presents significant challenges. To tackle this, we propose GANESH, a novel framework that refines lensless captures while simultaneously rendering novel views. Existing 2D approaches address this task in a sequential, two-step process, resulting in suboptimal 3D reconstruction quality. In contrast, GANESH integrates these two stages into a unified framework, enabling joint optimization for superior novel view synthesis.
  }
  \label{fig:teaser}
\end{figure}

Numerous studies have investigated the problem of scene reconstruction from single-image lensless captures \cite{khan2020flatnet, bagadthey2022flatnet3d, ge2024lpsnet, rego2021robust}. For instance, FlatNet \cite{khan2020flatnet} employs a two-step process to recover the scene. Initially, a trainable inversion module is utilized to reconstruct most of the scene's details; however, the resulting output still contains significant noise, which is subsequently addressed by a refinement network. While there are several works in this field of 2D scene reconstruction, there has been little to no work in reconstructing a 3D scene from multi-view lensless captures. This advancement is particularly significant for applications such as endoscopic surgery, where the compact size of lensless cameras offers a substantial advantage. Achieving 3D reconstruction from multi-view lensless images could greatly benefit medical fields and AR/VR applications\cite{khan2023designing, guerit2021compressive}

Recently, NeRFs have amassed a lot of attention for their ability to reconstruct 3D scenes from real-world multi-view captures. Most NeRF-based methods rely on RGB images to model the underlying 3D scene. There have been several works in this space which prompt NeRFs with different image modalities like Thermal Images\cite{ye2024thermal}, Event Data\cite{hwang2023ev}, Multi-spectral captures\cite{zhu2023x}, Single-Photon Data\cite{jungerman2024radiance}. For instance, in Thermal NeRF\cite{ye2024thermal}, they present an approach for reconstructing novel views exclusively from thermal imagery, particularly beneficial in visually degraded robotics scenarios. Ev-NeRF\cite{hwang2023ev} learns to reconstruct multi-view images from a raw stream of event data captured by a neuromorphic camera, which helps in better reconstruction, especially in high dynamic range scenes. Despite these promising developments, a significant limitation of current NeRF models is their reliance on scene-specific training. 

One might consider using established refinement techniques such as FlatNet \cite{khan2020flatnet} and then feeding the outputs to rendering networks like NeRF or Gaussian Splatting \cite{kerbl3Dgaussians}. Though this is a viable option, it comes with its downside. NeRF and Gaussian Splatting operate on the principle of using images solely for supervisory purposes rather than as direct inputs. Such models cannot be trained on paired multi-view lensless and RGB images, resulting in poor novel view synthesis quality. Secondly, while Gaussian Splatting offers improved computational efficiency compared to NeRF, it may be susceptible to overfitting the noisy outputs produced by FlatNet, potentially resulting in suboptimal reconstruction quality. Finally, each model must be re-trained from scratch whenever presented with a new set of images, limiting their scalability and practical application across diverse scenarios.

Generalizable Radiance Fields methods have recently gained traction due to their ability to perform on-the-fly inference on new scenes without specific training. Many of these approaches \cite{wang2021ibrnet, gnt, zhu2023caesarnerf} utilize a set of source views and enforce epipolar constraints across them to generate novel target views. The current state-of-the-art method, GNT \cite{gnt}, employs a transformer-based architecture to aggregate epipolar information from multiple views effectively. It then accumulates these point features along each ray to compute the final pixel color, enabling accurate and efficient rendering of new views. However, as previously noted, most Radiance Fields methods have predominantly focused on using RGB images as input, with limited exploration of alternative modalities such as lensless captures. Given the diverse applications of lensless imaging, incorporating this modality into radiance fields presents significant opportunities for expanding their utility across various fields.

In this paper, we introduce a novel methodology that enables us to reconstruct scenes from multi-view lensless captures in a generalizable setting. Unlike traditional methods that require scene-specific training for each new dataset, our technique can generalize across various multi-view lensless inputs to render novel views. Our proposed method, \textbf{GANESH}, can effectively reconstruct 3D scenes from lensless data. Our model is trained on extensive data of synthetically generated multi-view lensless images. Despite being trained exclusively on synthetic data, the model can refine and render novel views when applied to real multi-view lensless captures. Experimental results illustrate the model’s effective generalization to both synthetic and real-world scenes.
Additionally, our method allows for scene-specific tuning with minimal finetuning steps, enhancing reconstruction quality. We also present \textit{LenslessScenes}, a dataset of real-world multi-view lensless captures comprising six distinct scenes. These scenes, acquired in a controlled laboratory setting, are accompanied by ground truth data for precise quantitative evaluation. The key contributions of our work are as follows:
\vspace{-3mm}
\begin{itemize}
    \setlength\itemsep{0mm}
    \item We present a novel framework that simultaneously achieves refinement and rendering of lensless captures. 
    \item Our approach is generalizable, i.e., it can render views on-the-fly without any need for scene-specific training.
    \item We present \textit{LenslessScenes}, the first dataset of multi-view lensless captures. 
    \item Our experimental results demonstrate that the proposed method outperforms existing techniques that separately handle refinement and novel view synthesis. 
\end{itemize}


\begin{figure*}
  \includegraphics[width=\textwidth]{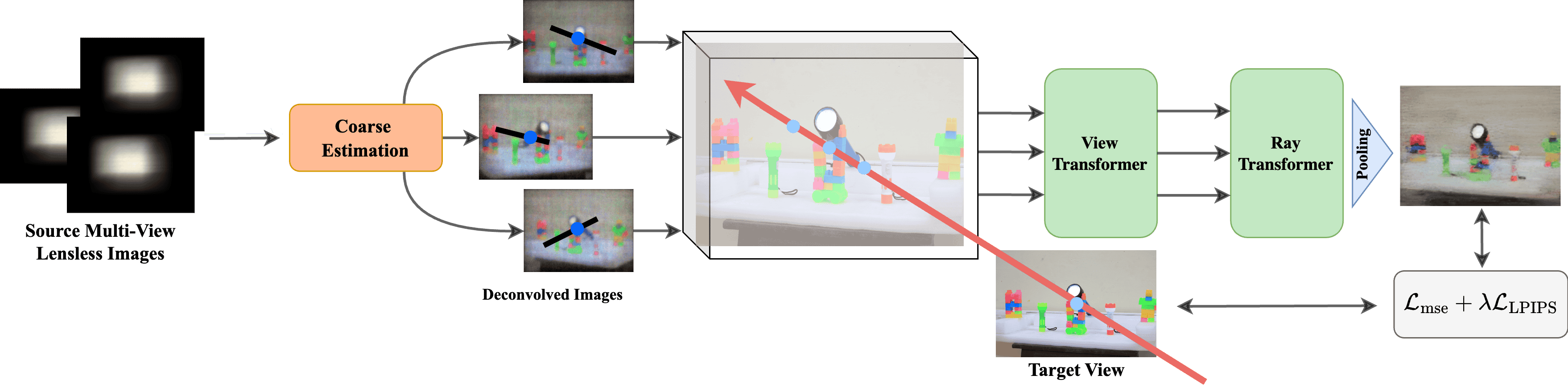}
  \caption{Overview of GANESH: 1) Given multi-view lensless images of a scene, we first Wiener deconvolve the lensless captures to obtain coarse images. 2) These are then passed onto a deep convolutional network to extract features for every input view. 3) Using the source view features, we estimate the target refined rendered view via an epipolar-based rendering pipeline. 4) By supervising this pipeline end-to-end on paired synthetic data, our model learns to inherently refine the coarse estimated images and simultaneously render novel views eliminating the need for a separate refiner. Our method can directly generalize to any new scene during inference. 
  }
  \label{fig:modelarch}
\end{figure*}

\section{Related Works}

\subsection{Lensless Imaging}
Lensless imaging refers to capturing images of a scene without using a traditional lens to focus incoming light. Historically, this technique has been widely utilized in X-ray and gamma-ray imaging for astronomical purposes \cite{dicke1968scatter, caroli1987coded}, but its application in the visible spectrum has only recently been explored. In lensless systems, the scene is captured either directly by the sensor \cite{kim2017lensless} or after being modulated by a mask element \cite{asif2016flatcam, antipa2017diffusercam, stork2013lensless}. Our research focuses specifically on mask-based lensless imaging, where replacing the lens with a mask leads to a highly multiplexed sensor capture that does not directly resemble the original scene. Consequently, advanced computational techniques are required to reconstruct the image. FlatNet \cite{khan2020flatnet} addresses this by employing a trainable inversion module coupled with a U-Net refinement architecture to recover the scene from a lensless capture. FlatNet3D \cite{bagadthey2022flatnet3d} further extends this by predicting the scene's intensity and depth from a single lensless capture using a neural network. However, no prior works have explored the use of multi-view images in lensless imaging. Our proposed method, GANESH, seeks to refine and render novel views from multi-view lensless captures, enabling simultaneous refinement and novel view synthesis.

\subsection{Neural Radiance Fields}
NeRF \cite{mildenhall2020nerf} utilizes a Multi-Layer Perceptron (MLP) to represent a scene as a continuous 5D function, incorporating both spatial location and viewing direction. This framework encodes the scene's geometry and appearance by mapping a 3D spatial function and a 2D directional function to outputs of color and density. Since its introduction, numerous works have sought to enhance NeRF's rendering capabilities\cite{park2021hypernerf, wizadwongsa2021nex, muller2022instant, chen2022tensorf}. Mip-NeRF\cite{barron2021mip, barron2022mip}, for instance, improves upon the original method by employing an approximate cone tracing approach rather than the ray tracing method used in standard NeRF. PointNeRF \cite{xu2022point} further advances this framework by introducing feature point clouds as an intermediate step in the volumetric rendering process, enhancing the overall quality of the rendered output.

NeRF has also been extended to non-RGB input modalities, such as in Thermal-NeRF \cite{ye2024thermal} and Hyperspectral NeRF \cite{chen2024hyperspectral} etc. Thermal-NeRF \cite{ye2024thermal} reconstructs 3D scenes using infrared (IR) images as input, focusing on preserving thermal characteristics more accurately. Similarly, Hyperspectral NeRF \cite{chen2024hyperspectral} adapts NeRF for hyperspectral imaging, which captures data across a broad range of the electromagnetic spectrum. In this work, we explore a related approach by reconstructing 3D scenes from lensless captures, leveraging this alternative imaging modality to extend NeRF’s capabilities.

\subsection{Generalizable Radiance Fields}

A significant drawback of NeRF is its lack of generalization, as each model is specifically trained for a single scene and cannot be easily transferred to new, unseen scenes. Various approaches, such as MVSNeRF \cite{chen2021mvsnerf}, IBRNet \cite{wang2021ibrnet}, and Generalizable NeRF Transformer (GNT) \cite{gnt}, have addressed this limitation by developing models capable of generalizing across different scenes. MVSNeRF\cite{chen2021mvsnerf} enhances novel view synthesis speed by incorporating multi-view stereo methods. IBRNet\cite{wang2021ibrnet} offers a generalizable image-based rendering framework that generates novel views from arbitrary inputs without requiring per-scene optimization. GNT\cite{gnt} leverages a transformer-based architecture to synthesize novel views by aggregating the information across source views based on epipolar constraints. Building on these advancements, we propose a generalizable model designed explicitly for novel view synthesis from lensless captures.


\section{Preliminary: Generalizable NeRF Transformer}
Our method is built on the Generalizable NeRF Transformer (GNT) framework, aimed at generating accurate images from noisy input views. Given $N$ calibrated source views with known pose information $\{\Mat{I}_{i}, \Mat{P}_{i}\}_{i=1}^{N}$, the objective is to synthesize a novel target view $\Mat{I}_{T}$, even for scenes not encountered during training, ensuring generalizability. To achieve this, deep convolutional features $\Mat{F}_{i}$ are extracted from each source view. During the rendering process, rays are cast into the scene, with $K$ points sampled along each ray, defined by the camera center $\Mat{o}$ and ray direction $\Mat{d}$. Each point is projected onto the source views using a projection operator $\Pi_{i}$, and the nearest features in the image plane are retrieved.

These features are combined into a point feature $\Mat{f}(t)$ using a permutation-invariant aggregation function $\mathcal{F}_{view}$ that is trained to handle occlusions. 
\begin{equation}
    \Mat{f}(t) = \mathcal{F}_{view}(\{\Mat{F}_{\Pi_{i}(\Mat{r}(t))}\}_{i=1}^{N})
\end{equation}

Finally, the accumulated point features are used to compute the target color $\Mat{c}(r)$ via an aggregation function $\mathcal{F}_{point}$, resulting in the rendered image.
\begin{equation}
    \Mat{c}(r) = \mathcal{F}_{point}(\{\Mat{f}(t_{i})\}_{i=1}^{K})
\end{equation}

We leverage GNT's capabilities to generate target novel views and visually enhance the reconstructions of lensless captures, avoiding the need for a refinement network before novel view synthesis.

\section{GANESH}
\textbf{Overview.} We introduce GANESH, a novel framework to perform generalizable novel view synthesis from lensless captures, as illustrated in Fig. \ref{fig:modelarch}. The task is to generate refined novel views from $N$ calibrated multi-view lensless images of a scene, with known camera poses, while ensuring the model generalizes to unseen scenes. Our method builds upon existing GNT architecture\cite{gnt} but conditions the scene representation and rendering processes based on the captured multi-view lensless images. First, these lensless captures are passed through a simple Wiener deconvolution filter to obtain a coarse estimate of the scene (Sec \ref{sec:wiener}). The deconvolved outputs of this filter are then passed on to a generalizable view synthesis model, which performs both refinement and rendering simultaneously. Such a pipeline can be trained end-to-end on synthetically generated scenes (Sec \ref{sec:simulation}) and directly transferred to any real scene without additional optimization (Sec. \ref{sec:training}).

\subsection{Coarse Scene Estimation}
\label{sec:wiener}
Given the global multiplexing of lensless captures, we cannot directly feed them into the radiance fields model to render novel views. Hence, to reconstruct the RGB image from the lensless captures, these need to be deconvolved with the lensless camera's point spread function (PSF) to obtain coarse reconstructed images. For this, we utilize wiener deconvolution, which accepts the lensless capture and the point spread function as the input and returns the reconstructed image. For an RGB image $I$ and a PSF kernel $H$, the observed lensless image is given by:
\begin{equation}
G(x,y) = (I * H)(x,y) + n(x,y)
\end{equation}
The wiener deconvolution will produce the following estimate for the RGB image:
\begin{equation}
    \hat{I}(\omega_x, \omega_y) = \frac{H^*(\omega_x, \omega_y)}{|H(\omega_x, \omega_y)|^2 + K} G(\omega_x, \omega_y),
\end{equation}
where $\hat{I}(\omega_x, \omega_y)$ is the Fourier transform of the estimated original image, 
$H(\omega_x, \omega_y)$ is the Fourier transform of the PSF, $H^*(\omega_x, \omega_y)$ is the complex conjugate of $H(\omega_x, \omega_y)$, $G(\omega_x, \omega_y)$ is the Fourier transform of the lensless capture, and $K$ is the noise-to-signal ratio.
Note that the deconvolved outputs are extremely noisy and require refinement to reconstruct the scene. 

\subsection{Simulating Lensless Imaging}
\label{sec:simulation}

Given the absence of a large-scale paired dataset comprising multi-view lensless captures and corresponding ground-truth RGB images, which is essential for training our generalizable model, we propose to simulate lensless data using existing multi-view RGB datasets. Specifically, we approximate lensless captures by convolving each ground-truth image with the lensless camera's point spread function (PSF). However, lensless measurements are frequently corrupted by noise in real-world applications, necessitating refinement steps to recover the original scene. We artificially introduce 40dB of Gaussian noise to the convolved lensless captures to simulate these practical conditions. This addition of noise ensures that the model is exposed to noisy data during training, helping it learn robust features that transfer well from synthetic to real-world scenarios.

An important design choice in our simulation process involves using a grayscale PSF map instead of an RGB PSF map for the convolution operation. Through empirical studies, we found that the grayscale PSF map more accurately mimics real-world lensless captures, which inherently lack color-channel-specific information due to the absence of a lens. As a result, the grayscale PSF map provides a more realistic approximation of the sensor measurements in lensless imaging systems, leading to improved reconstruction quality during inference, as demonstrated in the ablation study presented in Sec. \ref{sec:ablation}. 

\begin{figure}
  \includegraphics[width=0.47\textwidth]{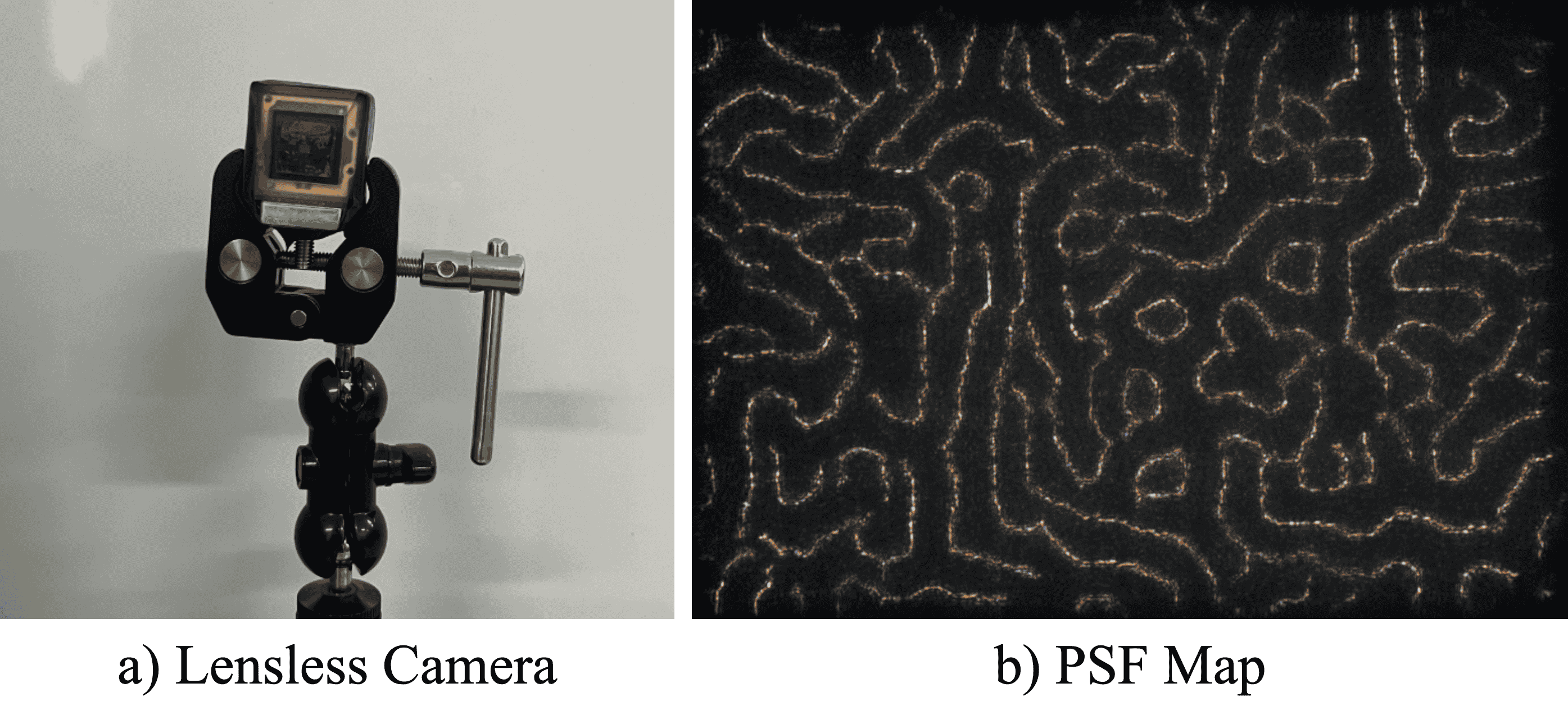}
  \caption{a) Lensless camera setup used to capture the real-world dataset \textit{LenslessScenes}. b) The calibrated Point Spread Function (PSF) is used for simulating lensless captures in our synthetically generated dataset.}
  \label{fig:setup}
\end{figure}

\begin{figure*}
  \includegraphics[width=\textwidth]{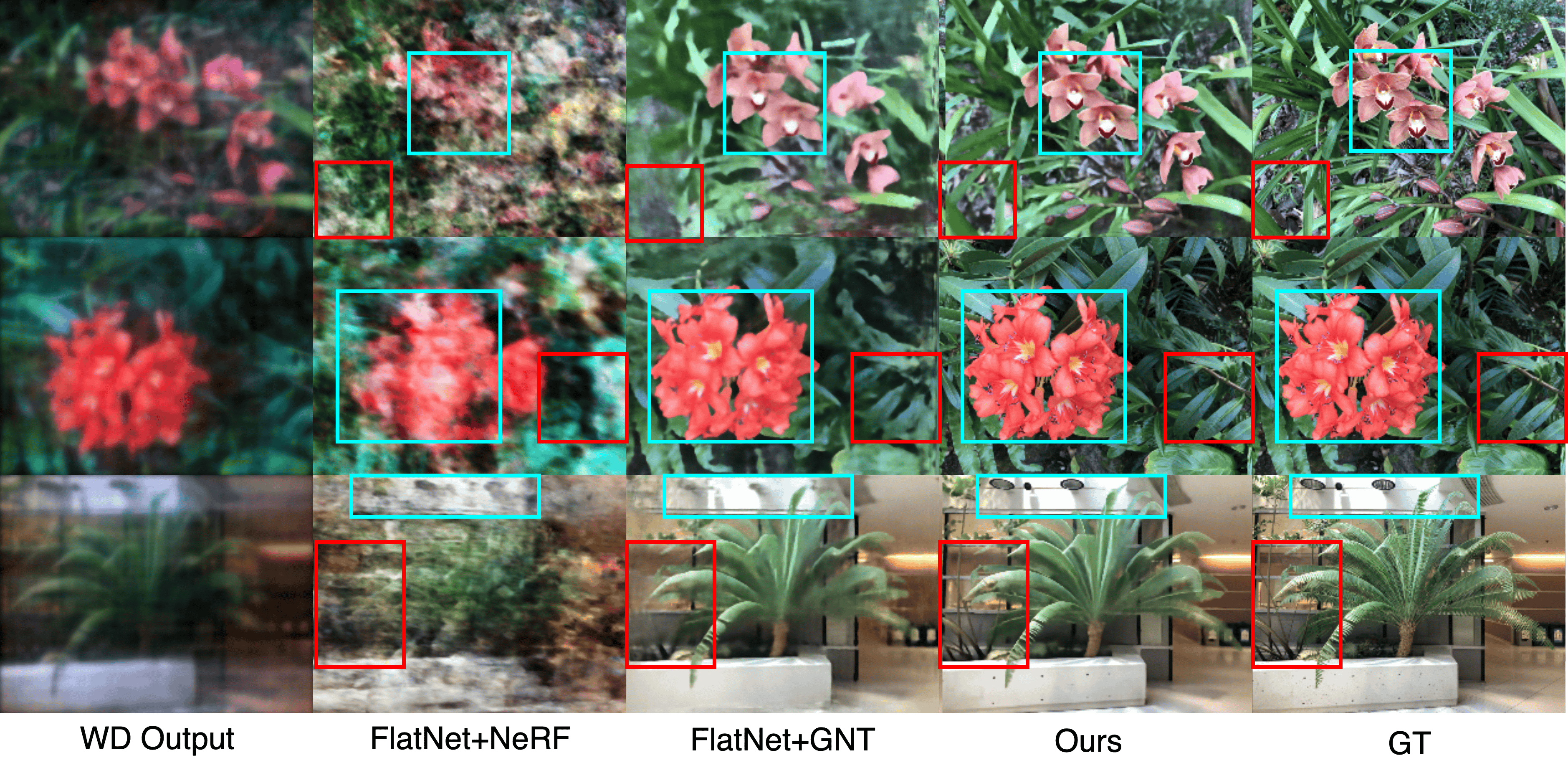}
  \caption{Qualitative results for scene-specific experiment on the synthetic NeRF-LLFF dataset. FlatNet+NeRF baseline exhibits significant artifacts and fails to preserve critical scene geometry. While FlatNet+GNT improves scene geometry reconstruction, it introduces excessive smoothing, resulting in the loss of high-frequency details. In contrast, our proposed method accurately reconstructs scene geometry and renders novel views, preserving high-frequency details and delivering superior visual fidelity. Note that the input to all the baselines and our model is the direct lensless capture. In the first column of this figure and all the subsequent figures, we show the Wiener Deconvolution (WD) output just for visualisation. 
  }
  \label{fig:scene-specific}
\end{figure*}

\subsection{Training and Inference}
\label{sec:training}

To optimize the network end-to-end, we use 2 losses to ensure view consistent rendering and accurate refinement. 

\textbf{Mean squared error:} We use MSE to measure the distortion between the ground truth and the rendered output given by.
\begin{equation}
    \mathcal{L}_{\text{MSE}} = \frac{1}{N} \sum_{i=1}^{N} (x_i - \hat{x}_i)^2
\end{equation}
where, \ensuremath{x} and \ensuremath{\hat{x}} represents the ground truth and the predicted images, respectively. 

\textbf{Perceptual Loss \cite{zhang2018unreasonable}:} In addition to MSE loss, we employ a perceptual loss to capture higher-level feature similarities between the ground truth and the rendered output. We use a pre-trained VGG-19 network to achieve this, extracting features from both the ground truth and predicted views at various layers. The perceptual loss is formulated as follows: 

\begin{equation}
    \mathcal{L}_{\text{Perceptual}} = \sum_{l} \lambda_l \frac{1}{N_l} \sum_{i,j} (F_l(x)_{i,j} - F_l(\hat{x})_{i,j})^2
\end{equation}
where \ensuremath{F} represents the VGG network and \ensuremath{\lambda_l} represents the weight given for each layer in the network.

\textbf{Final Loss. }
A weighted sum of the MSE and Perceptual Loss is taken to compute the final loss. 
\begin{equation}
    \mathcal{L} = \mathcal{L}_{\text{mse}} + \lambda \mathcal{L}_{\text{perceptual}}
\end{equation}
Through training on synthetically generated lensless scenes, we observe that GANESH successfully generalizes to real-world data without additional finetuning. This generalization capability aligns with results observed in prior 2D-based methods \cite{khan2020flatnet, bagadthey2022flatnet3d}, extending the hypothesis into the 3D domain and confirming its applicability to lensless image reconstruction.

\begin{figure*}
  \includegraphics[width=\textwidth]{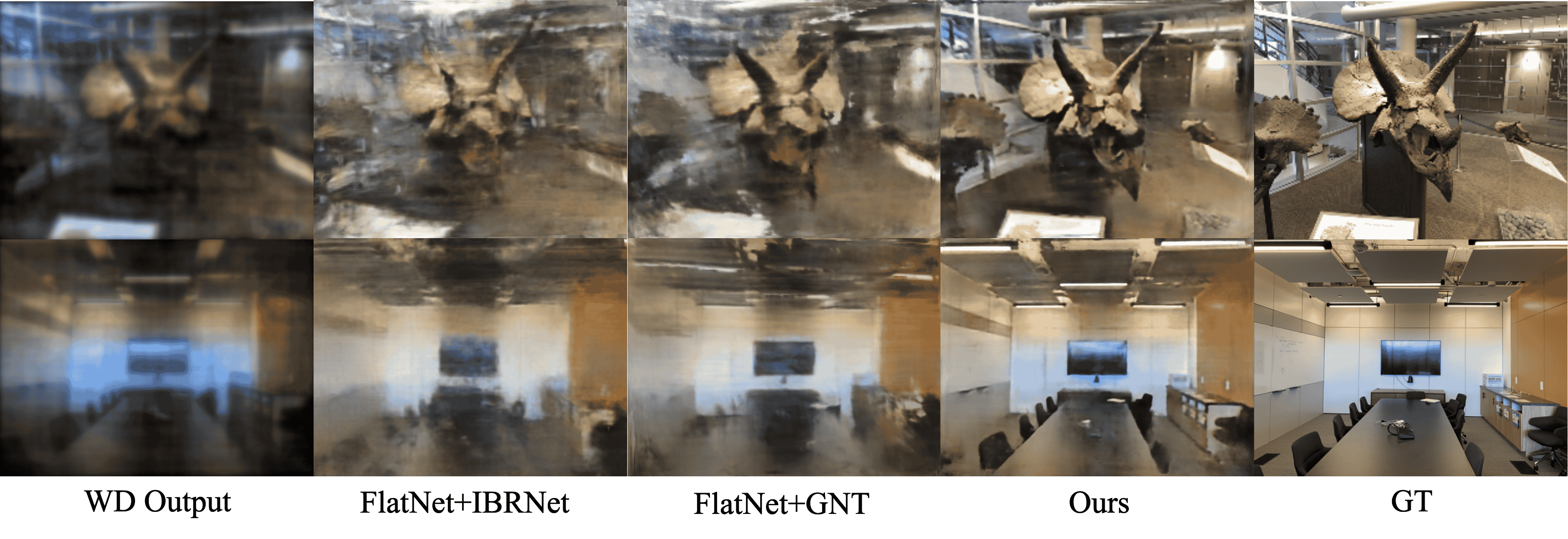}
  \caption{Qualitative results for generalizable setting conducted on the synthetic NeRF-LLFF dataset. We observe that the FlatNet+IBRNet and FlatNet+GNT baselines fall short in rendering high-fidelity novel views compared to our method. Our approach demonstrates superior recovery of fine geometry and textures. 
  }
  \label{fig:generalisable}
\end{figure*}

\section{Experiments and Results}
\label{sec:formatting}

\subsection{Datasets}
We leverage the IBRNet~\cite{wang2021ibrnet} and LLFF~\cite{mildenhall2019llff} datasets to create a synthetic dataset comprising lensless images and their corresponding ground truth RGB images. These datasets, which are well-established benchmarks for novel view synthesis (NVS), include a total of 110 scenes. For validation purposes on synthetic scenes, we utilize the NeRF-LLFF dataset~\cite{mildenhall2020nerf}.

\subsection{LenslessScenes Real-World Dataset}
To complement the synthetic data and test our model's robustness, we collected the first real-world multi-view lensless dataset. A set of 7 scenes was collected in the lab environment, consisting of an average of around 20 frames, each collected in a forward-facing setting. 
We replicate the setup of FlatNet\cite{khan2020flatnet} to collect real-data captures along with ground truth labels using a monitor capture setup. We used the \textit{BASLER Ace acA4024-29uc} lensless camera to capture the scenes, see Fig. \ref{fig:setup}. This data collection involved multiple scenes, each captured with a detailed environmental setup. A point-sized light source was employed to calibrate the camera’s point spread function (PSF), which is essential for accurate reconstruction. Additionally, a white display screen was used to capture environmental noise, which was subsequently subtracted from the images to enhance the quality of the captured data.

\subsection{Implementation Details}
Our entire pipeline is trained end-to-end using datasets of multi-view posed images. For consistency, we adopt the same input view sampling strategy employed by IBRNet~\cite{wang2021ibrnet}, selecting 8 to 12 source views during training while fixing the number of source views to 10 during inference for novel scenes. Instead of training the model from scratch, we initialize our network with a pretrained checkpoint from GNT~\cite{gnt}, allowing us to leverage its generalization capabilities.

The optimization of our model for rendering clean images is carried out using the Adam optimizer\cite{kingma2014adam}, with an initial learning rate set to $5\times10^{-4}$, which decays progressively over 300k training iterations. In each iteration, 576 rays are cast, with each ray sampling 192 points. The weight $\lambda$ in our loss function is assigned a value of 0.4, while the parameter $K$ from the Wiener Deconvolution process is 0.00045. All experiments are conducted on a single NVIDIA RTX 3090 GPU, with the entire training process taking approximately 24 hours to complete. As we cannot run COLMAP\cite{schonberger2016structure} directly on the lensless captures, we use the images recovered from FlatNet to run COLMAP and extract camera poses and bounds.  

\subsection{Comparisons}
In the absence of prior research specifically addressing novel view synthesis for lensless imagery, we propose several baseline approaches to evaluate our method.

\textbf{FlatNet+NeRF.} This approach involves first applying FlatNet to refine lensless captures, followed by utilizing NeRF for rendering. This is a major downside of using scene-specific methods like NeRFs since they rely on the supervision of images and hence cannot be trained explicitly to refine the lensless captures. Additionally, this baseline does not offer generalization across different scenes.

\textbf{FlatNet+IBRNet.} Here, we replace NeRF with the generalizable IBRNet for rendering, while maintaining FlatNet as the refinement module.

\textbf{FlatNet+GNT.} This baseline adopts a similar strategy to the previous ones but uses GNT instead of IBRNet for rendering. Both FlatNet+IBRNet and FlatNet+GNT are designed to generalize across different scenes.

\begin{table}[H]
  \centering
  \caption{
  Quantitative results for scene-specific experiment on synthetic dataset averaged across the 8 scenes from the NeRF-LLFF dataset. The \sethlcolor{black!30}\hl{best} scores and \sethlcolor{black!15}\hl{second best} scores are highlighted with their respective colors}

  \resizebox{6cm}{!}{\begin{tabular}{lccc}
    \toprule
    Models 
     & \hspace{-0.2em}PSNR$\uparrow$\hspace{-0.2em} & SSIM$\uparrow$\hspace{-0.2em} & LPIPS$\downarrow$\hspace{0.2em}\\
    \midrule
    FlatNet+NeRF & 13.7 & 0.057 & 0.705\\ 
    FlatNet+GNT & \cellcolor{black!15}20.2 & \cellcolor{black!15}0.27 & \cellcolor{black!15}0.59 \\  
    \midrule
    Ours & \cellcolor{black!30}22.8 & \cellcolor{black!30}0.71 & \cellcolor{black!30}0.27\\ 
    \bottomrule
  \end{tabular}}
  \label{tab:scenespecific}
\end{table}

\begin{figure*}
  \includegraphics[width=\textwidth]{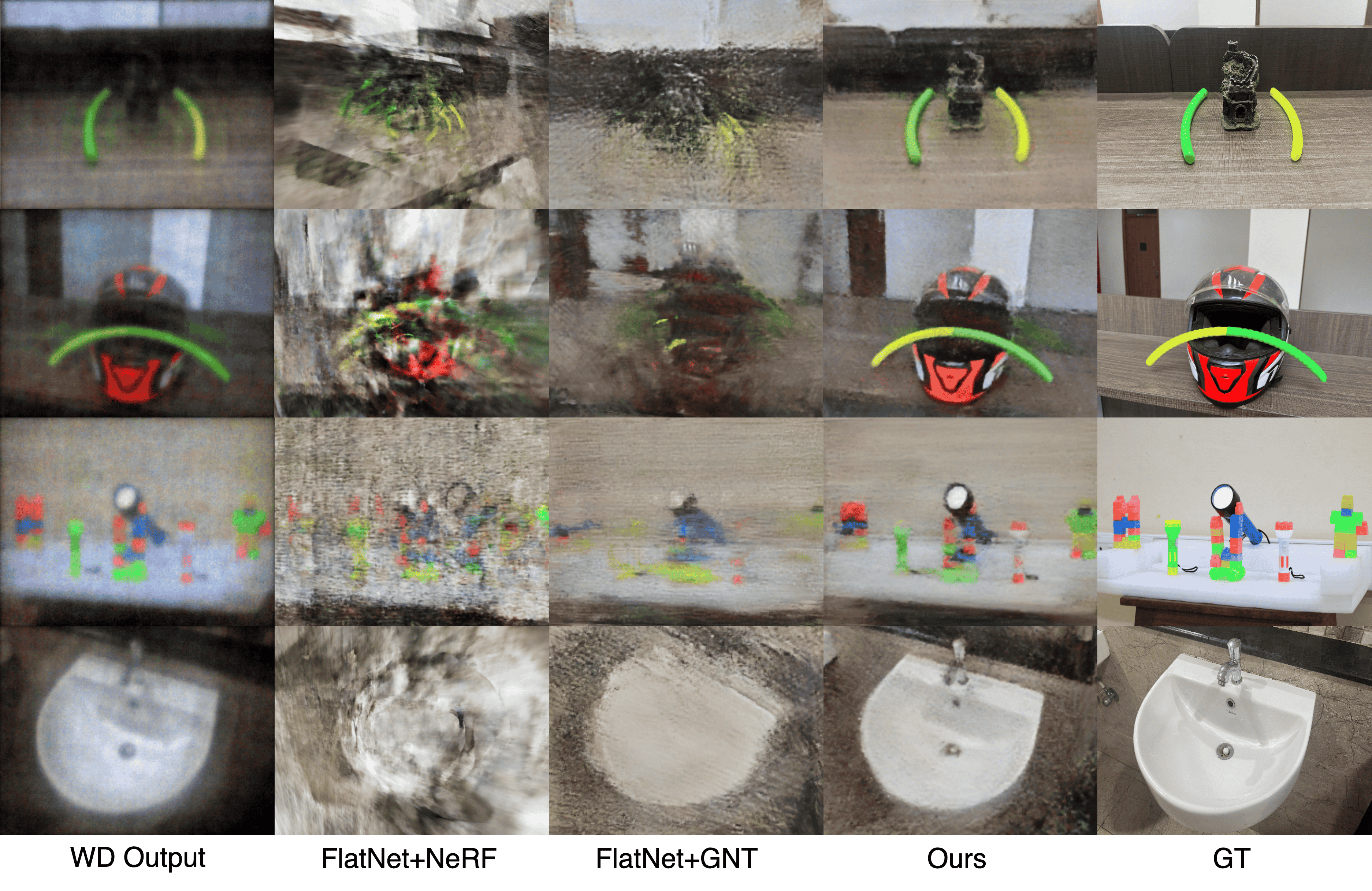}
  \caption{Qualitative results on the real-world LenslessScenes dataset. We show results for 4 scenes from the LenslessScenes dataset. Even though our model was trained on synthetic data, it learns to generalize to the real world captures and outperforms both the baselines in terms of render quality. 
  }
  \label{fig:realworld}
\end{figure*}

\subsection{Results}
\textbf{Scene-Specific Finetuning. } 
We aim to synthesize refined novel views from multi-view lensless images by training the model with supervision from corresponding ground truth RGB images. To evaluate the effectiveness of our approach, we compare the results with two baseline methods: FlatNet+NeRF and FlatNet+GNT. We supervise the NeRF model using FlatNet outputs as it is not possible to supervise it with ground truth labels since it takes in coordinates and viewing directions as input rather than an image. In contrast, for the FlatNet+GNT baseline, the outputs of FlatNet are provided as input to GNT, and supervision is conducted using ground truth RGB images. The evaluation is carried out on the synthetic NeRF-LLFF dataset, with the results detailed in Table \ref{tab:scenespecific}. Our proposed method, GANESH, demonstrates superior performance across all three metrics compared to both baseline models. This improvement can be attributed to the joint refinement and rendering strategy, which enhances the overall reconstruction quality. In a multi-view setup like ours, information lost in one view can be recovered from another, a feature that our method leverages. In contrast, the baseline methods perform refinement and rendering sequentially, missing out on the potential benefits of joint optimization.

Figure \ref{fig:scene-specific} illustrates the qualitative differences in performance. The FlatNet+NeRF model suffers from prominent ghosting artifacts, likely due to inconsistencies in the outputs generated by FlatNet, which are used for supervising NeRF. While FlatNet+GNT improves upon these issues through its more complex architecture, it still exhibits excessive smoothening effects. In contrast, our method achieves superior refinement and rendering, producing high-quality novel views. This demonstrates that our joint approach to refining and rendering in a 3D context significantly enhances the accuracy of novel view synthesis in lensless imaging.

\begin{table}[H]
  \centering
  \caption{
  Quantitative results for generalizable setting on synthetic NeRF-LLFF dataset.}

  \resizebox{6cm}{!}{\begin{tabular}{lccc}
    \toprule
    Models 
     & \hspace{-0.2em}PSNR$\uparrow$\hspace{-0.2em} & SSIM$\uparrow$\hspace{-0.2em} & LPIPS$\downarrow$\hspace{0.2em}\\
    \midrule
    FlatNet+IBRNet & 15.31 & 0.42 & 0.645\\ 
    FlatNet+GNT & \cellcolor{black!15}16.47 & \cellcolor{black!15}0.43 & \cellcolor{black!15}0.63 \\  
    \midrule
    Ours & \cellcolor{black!30}21.75 & \cellcolor{black!30}0.68 & \cellcolor{black!30}0.366\\ 
    \bottomrule
  \end{tabular}}
  
  \label{tab:generalisable}
\end{table}

\textbf{Generalizable Setting. } We evaluate our approach in a generalization scenario where the model is tested on unseen scenes, using eight scenes from the NeRF-LLFF dataset. The average results across these scenes are summarized in Table \ref{tab:generalisable}. Our method demonstrates clear superiority over both the FlatNet+IBRNet and FlatNet+GNT baselines, achieving higher performance across all three evaluation metrics. Figure \ref{fig:generalisable} provides a visual comparison of the novel views rendered by our method and the baselines. Our model successfully recovers intricate scene details, such as the subtle grooves on the plastic fortress and the veins on the leaves, with significantly higher fidelity than the other approaches. This highlights the effectiveness of our method in rendering high-quality novel views, even in challenging generalization settings.

\begin{table}[H]
  \centering
  \caption{
  Quantitative results on the real-world \textit{LenslessScenes} dataset. Our method shows strong generalisable capability across real scenes compared to the other baselines. }
  \resizebox{6cm}{!}{\begin{tabular}{lccc}
    \toprule
    Models 
     & \hspace{-0.2em}PSNR$\uparrow$\hspace{-0.2em} & SSIM$\uparrow$\hspace{-0.2em} & LPIPS$\downarrow$\hspace{0.2em}\\
    \midrule
    FlatNet+NeRF & 11.2 & 0.34 & 0.67\\ 
    FlatNet+GNT & \cellcolor{black!15}13.56 & \cellcolor{black!15}0.44 & \cellcolor{black!15}0.60 \\  
    \midrule
    Ours & \cellcolor{black!30}18.45 & \cellcolor{black!30}0.62 & \cellcolor{black!30}0.47\\ 
    \bottomrule
  \end{tabular}}
  \label{tab:realworld}
\end{table}

\textbf{Quantitative results on Real-world dataset} To assess the robustness of our model on real-world data, we evaluate its performance on the \textit{LenslessScenes} dataset. We compare the results against the FlatNet+NeRF and FlatNet+GNT baselines. As in previous experiments, the FlatNet+NeRF baseline supervises NeRF using the output from FlatNet, while FlatNet+GNT is applied directly to the real scenes without any finetuning for the real-world data. Quantitative and qualitative evaluations of these comparisons are presented in Table \ref{tab:realworld} and Fig. \ref{fig:realworld}. Qualitatively, our model demonstrates a superior ability to recover fine scene details compared to the baselines. For example, the shape and geometry of the toys surrounding the torch in the figure are visible, contrasting with the results of the baseline methods, where they are hardly discernable. Quantitative results further support this observation, showcasing improved performance when transferring from synthetic to real-world data. These findings highlight the efficacy of our joint refinement and rendering approach, which significantly enhances 3D scene reconstruction compared to methods that treat refinement and rendering as independent tasks.

\begin{minipage}[c]{0.22\textwidth}
\centering
\resizebox{3.5cm}{!}{\begin{tabular}{lc}
    \toprule
    Models 
     & \hspace{-0.2em}PSNR$\uparrow$/SSIM$\uparrow$/LPIPS$\downarrow$\\
    \midrule
    
    RGB PSF & 22.43/0.62/0.47\\ 
    Gray PSF & \cellcolor{black!30}22.75/\cellcolor{black!30}0.63/\cellcolor{black!30}0.39\\ 
    \bottomrule
  \end{tabular}}
\centering
\captionof{table}{Ablation Study: Grayscale vs RGB PSF.}
\label{tab:ablationpsf}
\end{minipage}
\begin{minipage}[c]{0.22\textwidth}
\centering
\resizebox{3.5cm}{!}{\begin{tabular}{lc}
    \toprule
    Models 
     & \hspace{-0.2em}PSNR$\uparrow$/SSIM$\uparrow$/LPIPS$\downarrow$\\
    \midrule
    Ours w/o noise & 11.2/0.34/0.67\\ 
    Ours with noise & \cellcolor{black!30}18.45/\cellcolor{black!30}0.62/\cellcolor{black!30}0.47\\ 
    \bottomrule
  \end{tabular}}
  
\centering
\captionof{table}{Ablation Study: Noise Augmented Training.}
\label{tab:ablationnoise}
\end{minipage}

\subsection{Ablation Studies}
\label{sec:ablation}
We conduct the following ablations to validate our lensless simulation pipeline and provide quantitative results evaluated on real world data to test the effectiveness of our simulation. 

\textbf{Grayscale vs RGB PSF map.} Our experimental results indicate that utilizing a grayscale PSF for reconstructing lensless images consistently outperforms using a 3-channel RGB PSF, as demonstrated in Table \ref{tab:ablationpsf}. We hypothesize that this advantage stems from the fact that the grayscale PSF more closely mirrors how actual lensless cameras capture images. Consequently, using the grayscale PSF map during the simulation process yields superior results when tested on to real-world datasets.

\textbf{Synthetic Noise.} While our model is resilient to low levels of noise, training with synthetic noise added to the lensless images proved crucial for adapting to real-world scenes. Gaussian noise was artificially introduced before reconstruction during the training on synthetic captures as read noise. We tested this model on real-world data, and the results are presented in Table \ref{tab:ablationnoise}. The significant noise observed in real-world captures necessitated this approach, and incorporating noise in the training pipeline enhanced the model's robustness and ability to generalize to real-world scenarios.

\section{Discussion}

\textbf{Limitations and Future Work. } While GANESH demonstrates the ability to refine and render novel views from lensless captures in both scene-specific and generalizable settings, it is not without limitations. A primary challenge it faces is that the model requires substantial training time when generalizing across diverse scenes, and its inference speed is not optimized for real-time applications, posing a limitation for on-the-fly rendering tasks. Finally, GANESH is an entirely data-driven model trained on an extensive dataset to mimic the reconstruction task. Integrating physical light transport models into radiance fields could be a promising direction for future improvements, combining data-driven approaches with physical principles for more accurate and efficient lensless rendering.

\textbf{Conclusions. } 
In this work, we present GANESH, a novel framework that integrates refinement and novel view rendering from multi-view lensless captures within a generalizable framework, demonstrating robustness in real-world scenarios. While existing approaches such as FlatNet for refinement and NeRF or Gaussian Splatting (GS) for rendering could be employed sequentially, they are fundamentally constrained by their reliance on image supervision, making training on extensive synthetic datasets impractical. In contrast, GANESH enables joint refinement and rendering, addressing this limitation and achieving superior performance in novel view synthesis. This approach is crucial for various applications, including medical imaging (e.g., endoscopy), augmented and virtual reality (AR/VR), and wearable technologies. 

{\small
\bibliographystyle{ieee_fullname}
\bibliography{main}
}

\newpage
\maketitle
\appendix

\section{Video Results}
This supplementary material comprises video results demonstrating results on both synthetic scenes and real-world scenes from the LenslessScenes dataset. We conduct a side-by-side comparison and show that our method produces view-consistent rendering and refinement compared to the baselines.  

\section{Analysis of PSF geometry}
\begin{figure*}
  \includegraphics[width=\textwidth]{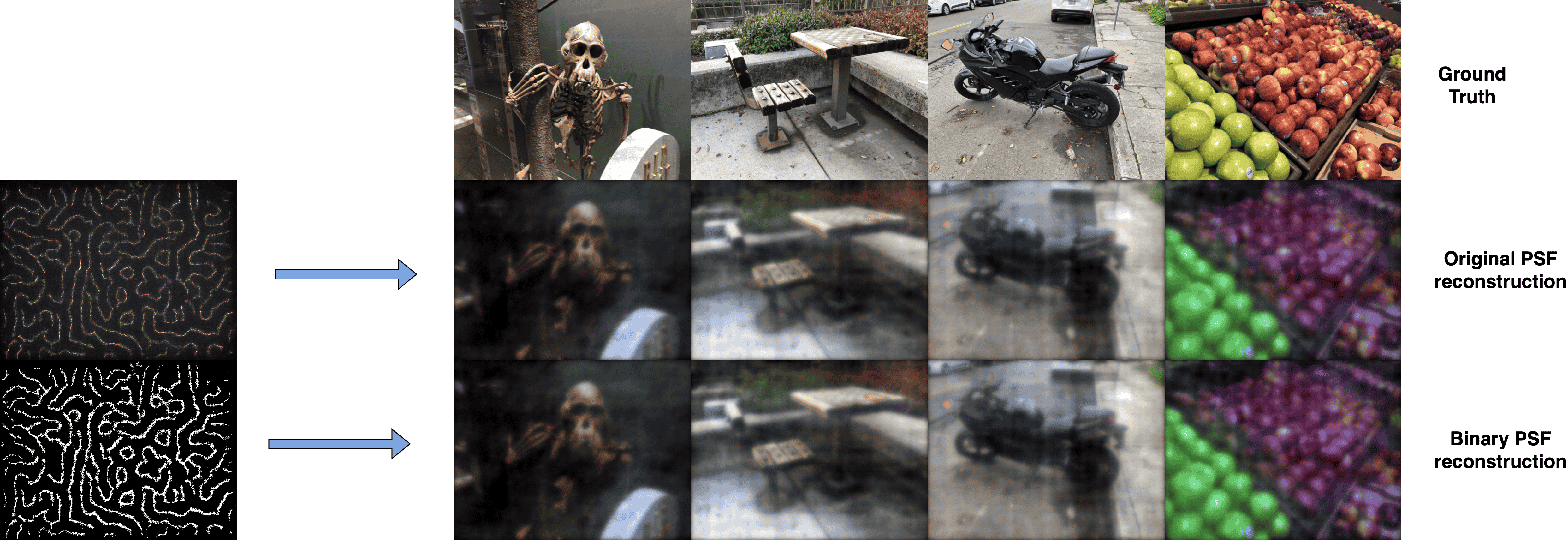}
  \caption{ The effect of Point Spread Function (PSF) on reconstructed images. This comparison showcases the output quality using two distinct PSF structures in our lensless multi-view reconstruction pipeline. Both PSF configurations yield reconstructed images of comparable quality.
  }
  \label{fig:colour}
\end{figure*}

\begin{figure*}
  \includegraphics[width=\textwidth]{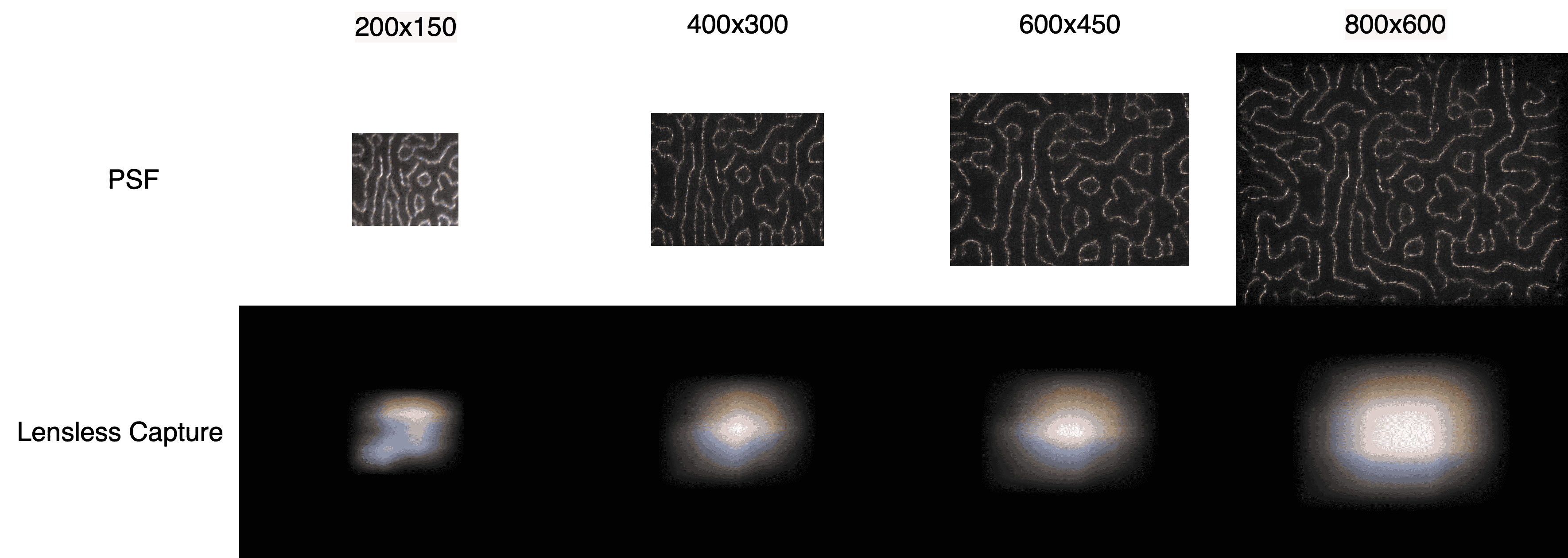}
  \caption{ Impact of Point Spread Function (PSF) size on image reconstruction quality. This figure illustrates the comparison between synthesized lensless captures after convolution with different PSF dimensions.
  }
  \label{fig:size}
\end{figure*}

In this section, we explore the impact of the structure and size of point spread functions (PSFs) on 3D reconstruction of lensless images. The observations on the effects of PSF structure and size are shown in Fig\ref{fig:colour} and Fig\ref{fig:size}, respectively.\\\\
\textbf{Structure.} We first analyzed the influence of PSF structure by comparing the results of PSFs in the RGB scale with those of binary PSFs, where only values of 0 and 1 were used. The binary PSF was derived by converting the original grayscale PSF into binary values, with thresholding applied to assign the binary values. Interestingly, the outputs produced by the RGB-scale PSF closely resembled those from the binary PSF, indicating that color information in the PSF has minimal impact on image reconstruction.This observation highlights that the structural characteristics of the PSF play a more significant role in decoding the image than the color properties. Thus, focusing on the PSF's shape, rather than its color composition, may lead to more effective results in the context of 3D image reconstruction.\\\\
\textbf{Size.} In addition to the PSF structure, we also studied the effects of PSF size. Smaller PSFs were generated by cropping the original PSF, and the resulting outputs were compared. These experiments provided insights into how varying PSF dimensions influence the quality of reconstructed images, highlighting importance of optimizing both PSF structure and size for improved lensless imaging performance. We cropped smaller samples from the original PSF, which had a size of 1518x2012. By examining PSFs of various dimensions, we observed notable differences in how image information was processed and reconstructed.\\\\
For the smaller PSFs, the distinct color components appeared as separate, clearly defined segments. These segmented regions indicate that, at smaller scales, the PSF’s resolution is sufficient to differentiate between various color channels, thereby maintaining a higher degree of color fidelity. However, as the PSF size increased, these color segments began to blend together, causing the distinct boundaries between colors to become less apparent. This merging effect suggests that larger PSFs smooth out fine details in the color channels, likely due to the increase in overlap between adjacent segments. Moreover, the increase in PSF size also led to more light being perceived in the resulting images. Larger PSFs gather more light, which enhances the brightness and overall intensity of the reconstructed images. However, this comes at the cost of reduced color separation and fine detail, as larger PSFs may introduce a form of blurring or averaging across the image.\\\\
These findings suggest that while larger PSFs can improve light sensitivity and overall image brightness, they may also compromise color accuracy and sharpness. On the other hand, smaller PSFs preserve finer details and color distinctions but may not capture as much light, potentially reducing image clarity in lower-light conditions. Therefore, an optimal balance between PSF size and structure must be carefully considered, depending on the specific requirements of the 3D reconstruction task, such as whether higher light sensitivity or color accuracy is prioritized.

\section{Synthetic Lensless Data Generation}
This section provides an overview of the experiments conducted to create synthetic lensless images. We utilized a calibrated point spread function (PSF) to convolve the images, effectively mimicking the characteristics of lensless image capture. We introduced up to 40dB of Gaussian noise $\Mat{n}$, to the images. 
\begin{equation}
I_{\text{noisy}}(x, y) = I(x, y) + n(x, y), \quad n(x, y) \sim \mathcal{N}(0, \sigma^2)
\end{equation} This process allowed us to generate synthetic lensless captures that closely approximate the output of actual lensless imaging systems.\\\\
To facilitate the reconstruction process, we implemented an initial coarse estimation of the RGB images using Wiener deconvolution. This step utilized the same PSF as the kernel, producing an intermediary reconstruction that serves as a starting point for our model. By providing this intermediate result, we enable our model to focus on refining the reconstruction while simultaneously synthesizing novel views.

\begin{figure*}
  \includegraphics[width=\textwidth]{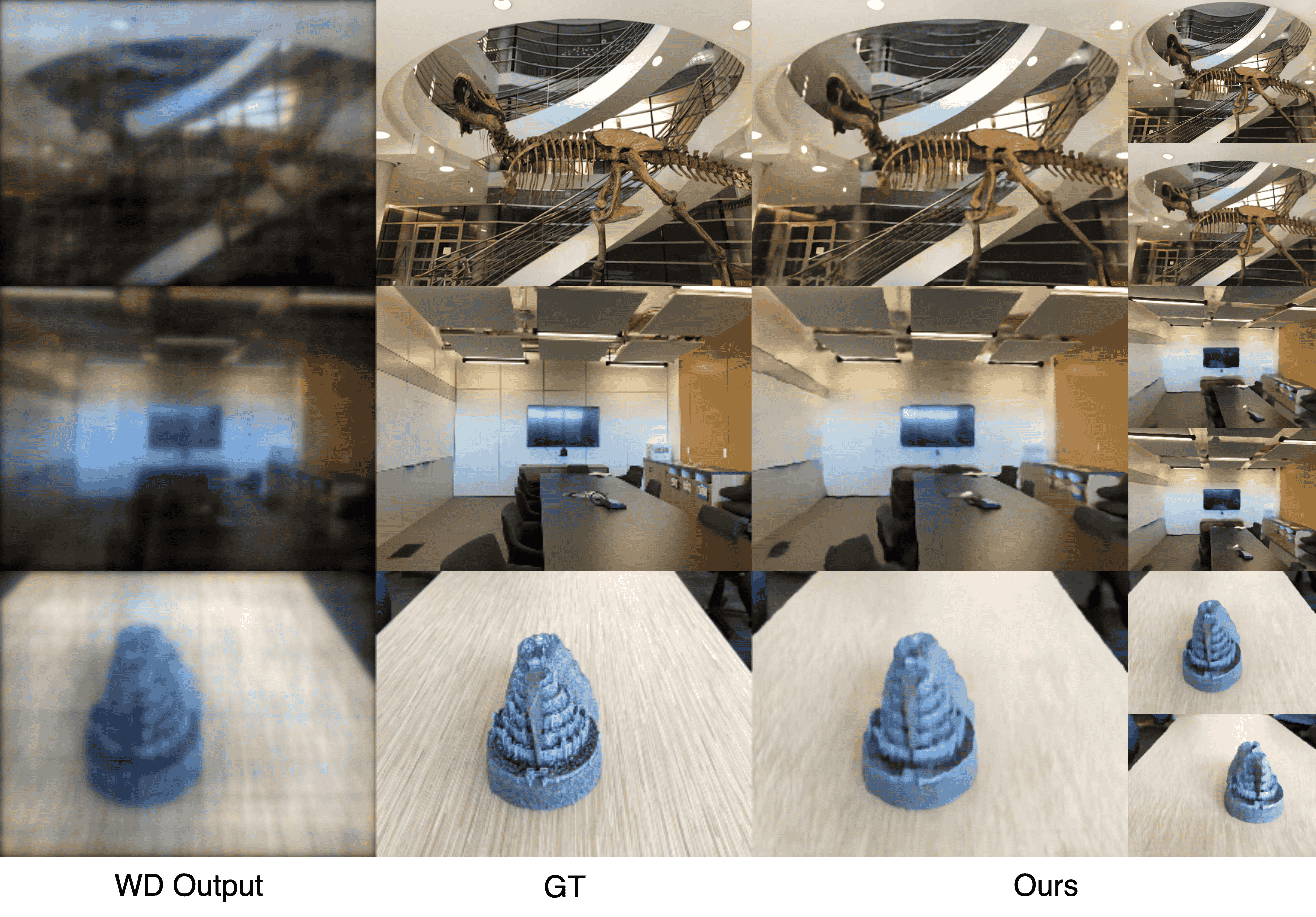}
  \caption{ Scene-specific performance of GANESH. This figure presents reconstructions from our model when trained on data from individual scenes. The images demonstrate the model's ability to capture detailed features and characteristics unique to each specific environment, illustrating the potential of tailored training approaches.
  }
  \label{fig:supplementary-scene}
\end{figure*}

\begin{figure*}
  \includegraphics[width=\textwidth]{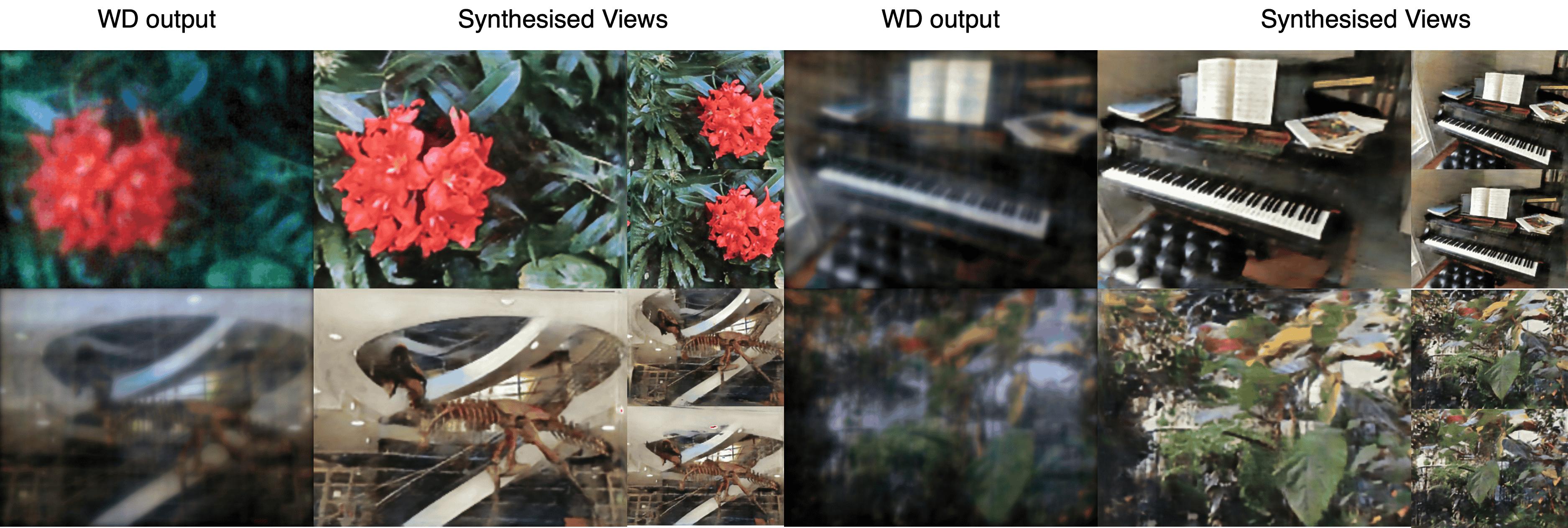}
  \caption{ GANESH results demonstrating consistent high-quality novel view synthesis. This figure presents a diverse array of scenes reconstructed by our GANESH model. The images showcase the model's ability to generate detailed and coherent novel views across various environmental conditions, object types, and spatial complexities, highlighting the consistency and versatility of our approach.
  }
  \label{fig:gallery}
\end{figure*}

\section{Results}

\begin{figure*}
  \includegraphics[width=\textwidth]{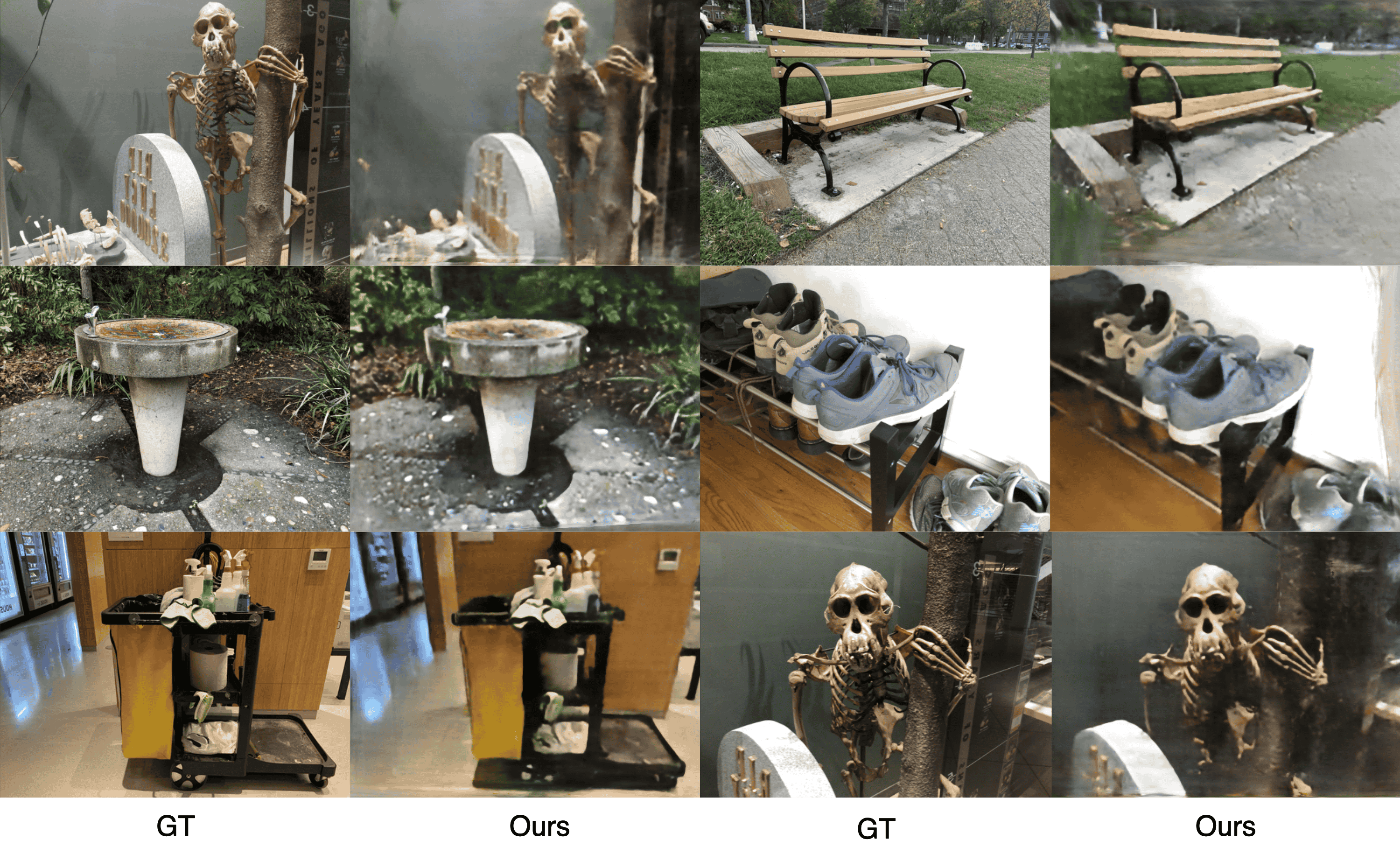}
  \caption{ Generalized performance of GANESH. This figure illustrates the model's performance when trained on a heterogeneous dataset encompassing various scenes and conditions. Observe the model's capacity to reconstruct diverse environments, demonstrating its versatility and robustness across different capture scenarios.
  }
  \label{fig:general}
\end{figure*}

Our experiments showcase the versatility and effectiveness of our approach across a wide range of scenes and capture conditions. Figure \ref{fig:supplementary-scene} presents a comprehensive set of results trained on scene-specific data. In scene-specific scenarios, our model demonstrates remarkably accurate reconstructions, preserving fine details particularly well in complex textures and intricate geometries. These results highlight the potential for highly tailored, high-fidelity reconstructions when the model is trained on data from a single scene. Figure \ref{fig:general} presents a comprehensive set of results trained on generalized data. The generalized model, while trained on a diverse dataset, shows robust performance across varied scenes it has not encountered during training. The generalization capability of GANESH is evident in its ability to handle different lighting conditions, object types, and spatial configurations. Both the scene-specific and generalized approaches produce visually compelling results. This dual capability underscores the flexibility of our method, making it suitable for a wide array of applications ranging from controlled environment captures to more challenging, diverse scene reconstructions.
\section{Multi-view rendering evenness}
In Figure \ref{fig:gallery}, we display the evenness in consistency of quality in the results of our model for multi-view scenes. We see that GANESH is capable of consistently rendering novel views from a variety of viewing angles, in a manner superior to other
baselines.

\section{Our Dataset}
\begin{figure*}
  \includegraphics[width=\textwidth]{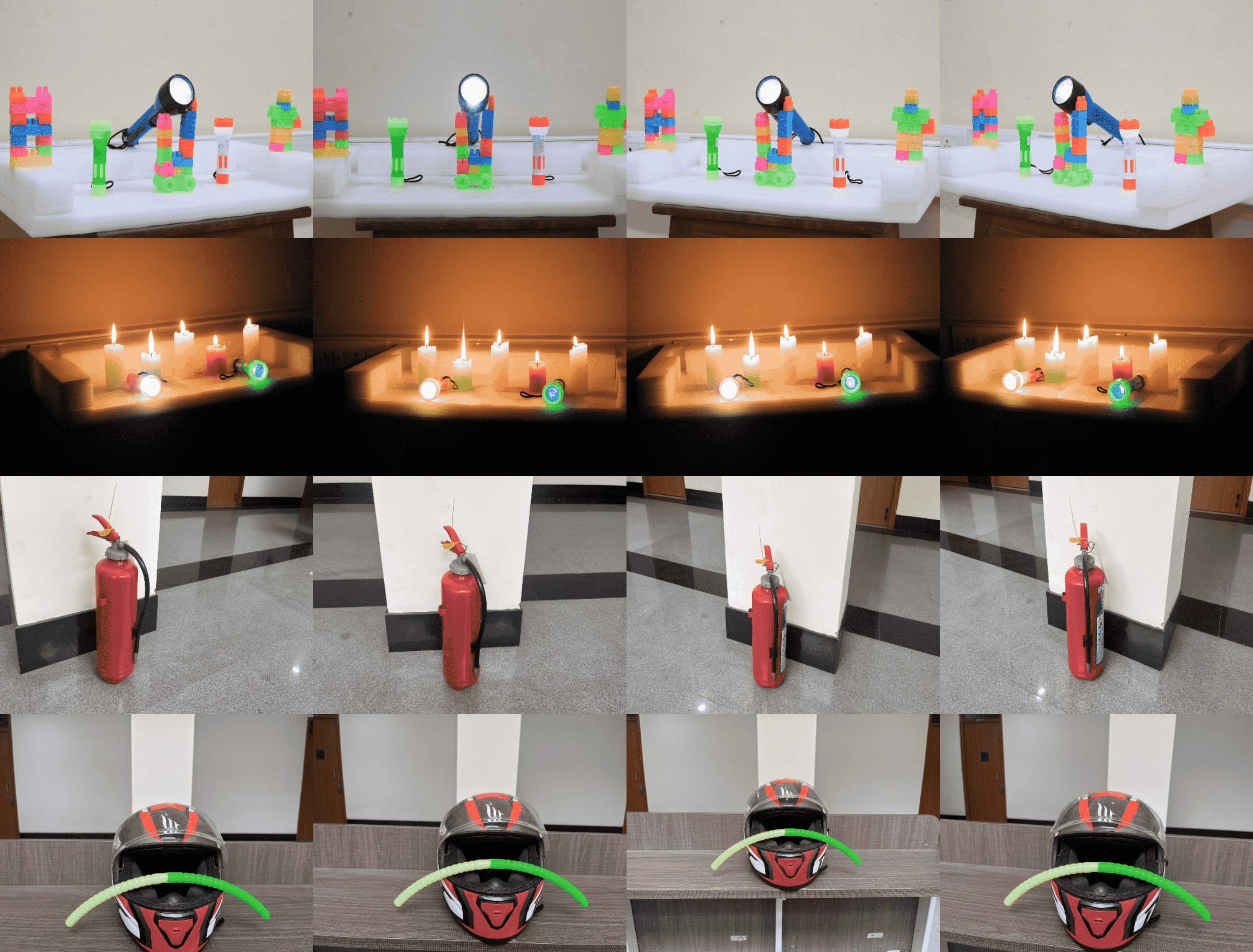}
  \caption{ Ground truth image captures for our multi-view lensless capture dataset LenslessScenes. The captures were taken in a controlled setting with pose and bound information calculated using COLMAP. Each scene has an average of around 20 views.
  }
  \label{fig:real_data}
\end{figure*}

In Figure \ref{fig:real_data}, we display a few ground truth views of our dataset \textbf{LenslessScenes}. Our dataset retains the format of the LLFF dataset, with scenes captured in a variety of lighting conditions and color palettes. In the \textbf{torch}, \textbf{toys} and \textbf{candles} scenes, luminous objects are included to test the capabilities of models in reconstructing areas of views that are affected by the bloom of the objects.

\end{document}